\pdfoutput=1

\documentclass[11pt]{article}

\usepackage[]{EMNLP2023}

\usepackage{times}
\usepackage{latexsym}
\usepackage{graphicx}

\usepackage[T1]{fontenc}

\usepackage[utf8]{inputenc}

\usepackage{microtype}

\usepackage{inconsolata}
\title{SentAlign: Accurate and Scalable Sentence Alignment}

\author{Steinþór Steingrímsson$^1$, Hrafn Loftsson$^1$ and Andy Way$^2$ \\
$^1$Department of Computer Science, Reykjavik University, Iceland  \\
$^2$ADAPT Centre, School of Computing, Dublin City University, Ireland \\
  \texttt{steinthor18@ru.is, hrafn@ru.is, andy.way@adaptcentre.ie}}

\begin{document}
\maketitle
\begin{abstract}
We present SentAlign, an accurate sentence alignment tool designed to handle very large parallel document pairs. Given user-defined parameters, the alignment algorithm evaluates all possible alignment paths in fairly large documents of thousands of sentences and uses a divide-and-conquer approach to align documents containing tens of thousands of sentences. The scoring function is based on LaBSE bilingual sentence representations. SentAlign outperforms five other sentence alignment tools when evaluated on two different evaluation sets, German--French and English--Icelandic, and on a downstream machine translation task.
\end{abstract}

\begin{figure}[b!]
\centering
  \centering
  \includegraphics[height=0.0825\textheight]{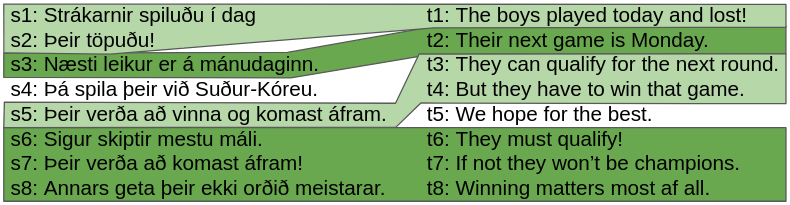}
  \caption{An automatic sentence alignment system aims to align source sentences $s_1,...,s_n$ with target sentences $t_1,...,t_n$ while using as few sentences as possible for each alignment. The figure shows examples of six alignment functions being applied while aligning eight sentences in Icelandic with eight sentences in English: 
  Contraction ($n$--$1$), expansion ($1$--$n$), deletion ($1$--$0$), insertion ($0$--$1$), substitution ($1$--$1$) and merging ($n$--$m$).}
  \label{fig:alignments}
\end{figure}

\section{Introduction}
\begin{figure}[b!]
\centering
  \centering
  \includegraphics[height=0.25\textheight]{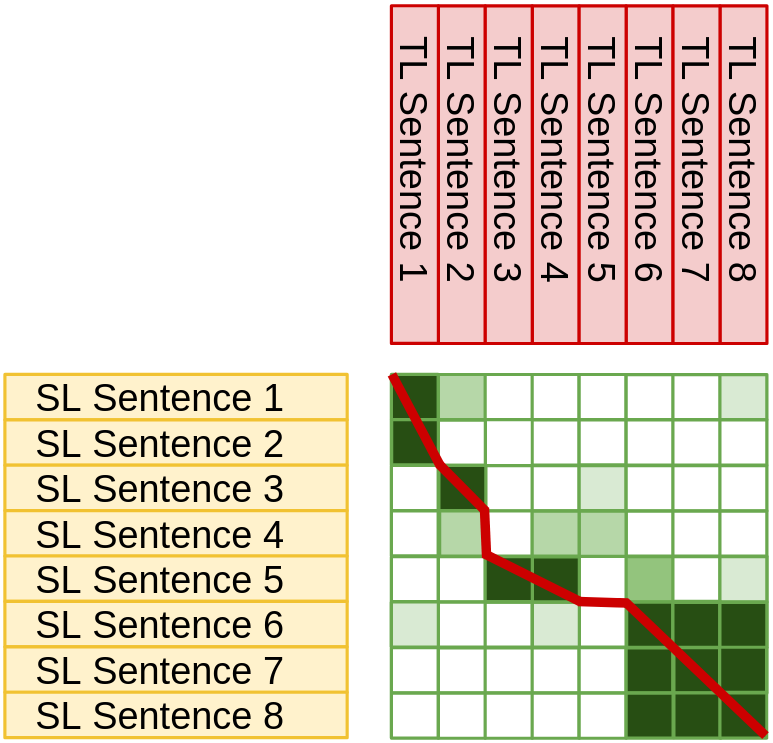}
  \caption{A two-dimensional alignment graph. The figure shows the path found through the graph after evaluating semantic similarity of all possible source (SL) and target language (TL) sentence pairs. Dark green nodes stand for the alignments selected by the system.}
  \label{fig:alignments_path}
\end{figure}

Sentence alignment is the task of finding matching sentences in two parallel documents, as illustrated in Figure \ref{fig:alignments}. It can be seen as a path-finding problem, with a list of source sentences on one axis in a two-dimensional graph and the target sentences on the other, as demonstrated in Figure \ref{fig:alignments_path}. Each potential sentence pair is represented by a node in the graph, or nodes when multiple sentences are grouped together. The nodes are assigned values using a scoring function. The objective of the sentence alignment algorithm is to find the optimal path through the graph. Typically, the path is continuous, although gaps may occur when one of the documents has sentences without corresponding counterparts in the other document. The alignments can also be non-monotonous, where sentences cross, resulting in differences in sentence order between languages. This  problem is often solved by chunking multiple sentences.

\begin{figure*}[t]
\centering
  \centering
  \includegraphics[height=0.16\textheight]{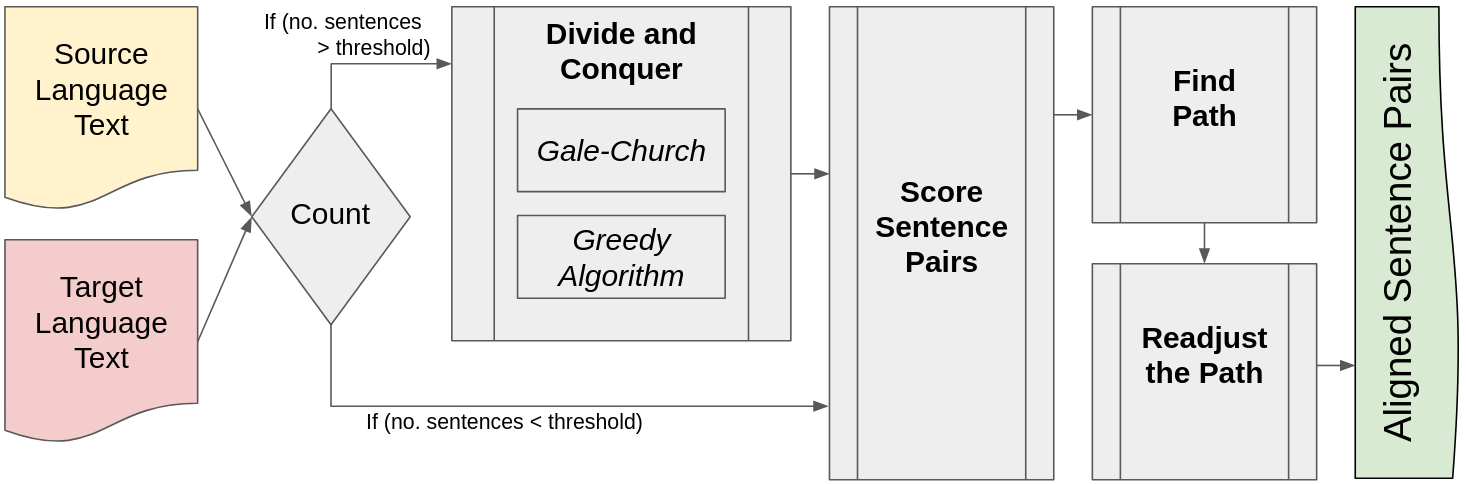}
  \caption{SentAlign Architecture.}
  \label{fig:sentalign_architecture}
\end{figure*}

Sentence alignment is a necessary processing step for parallel corpora to be useful for machine translation (MT). Neural machine translation (NMT) has been shown to be sensitive to misaligned training data (e.g. \citet{khayrallah-koehn-2018-impact}) so an accurate sentence aligner is highly important for NMT to unleash the full potential of the parallel corpora it is trained on. 


In this paper, we present SentAlign,\footnote{\url{https://github.com/steinst/sentalign/}} a sentence aligner with a user-friendly command line interface, able to align very large documents. As shown in Section \ref{sec:Evaluation} it outperforms other available sentence aligners when evaluated on a common evaluation set, as well as on a downstream MT task. SentAlign evaluates all possible alignment paths in fairly large documents, with up to a few thousand sentences in each language, and activates a divide-and-conquer (DaC) approach to reduce running time when the number of sentences exceed a user-defined threshold. To identify matching sentences in two languages, SentAlign applies a scoring mechanism based on LaBSE \cite{feng-etal-2022-language}, a model trained and optimized to produce similar representations for bilingual sentence pairs. The model, which employs both a masked language model \cite{devlin-etal-2019-bert} and a translation language model \cite{Lample2019CrosslingualLM}, is pre-trained on monolingual and bilingual data in 109 languages.

\section{Related Work}

\citet{gale1991} found that ``the correlation between the length of a paragraph in characters and the length of its translation was extremely high''. Motivated by that, they describe a method for aligning sentences based on a simple statistical model of character lengths. 

The similarity score for Hunalign \cite{varga_parallel} has two main components: token-based and length-based. The token-based component searches for shared words in the two sentences, using an automatically generated lexicon or an external one.
The length-based component is based on the ratio of longer to shorter sentences.
The similarity score is calculated for every sentence pair in the neighbourhood of the diagonal of the alignment graph.
Finally, a post-processing step iteratively merges 1--$n$ ($n>1$) and 0--1 segments wherever the resulting new segment has a better character-length ratio than the starting one. 

Gargantua \cite{braune-fraser-2010-improved} 
uses a two-step clustering approach to sentence alignment. It aims to find 1--$n$ and $n$--1 alignments, but does not search for many-to-many alignments. It uses sentence length-based statistics considering relative lengths in comparison to the mean length of source and target sentences, and translation likelihoods of each target word with all source words, according to IBM Model-1 \cite{brown-etal-1990-statistical}. 
It starts by looking for optimal alignments through the alignment matrix consisting only of 0--1, 1--0 and 1--1 correspondences. 
In a second step, the previously acquired alignments are merged into clusters containing up to $R$ sentences (4 by default) on either the source or target size, and if the merge produces a better score it is accepted. The final alignments are found when an optimal score has been obtained for the whole graph.

\begin{figure*}[t]
\centering
  \centering
  \includegraphics[height=0.12\textheight]{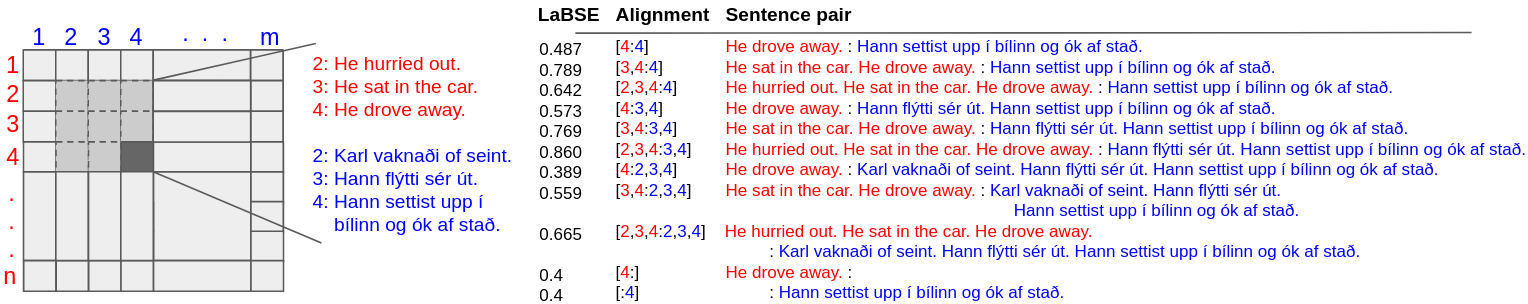}
  \caption{SentAlign searches for the best alignment that ends in node [4:4], with a maximum of 3 sentences merging on either side. LaBSE score is calculated for each alignment candidate. For insertions and deletions, where a sentence on either side is discarded, we assign the minimum threshold score, $S_{min}$.}
  \label{fig:sentalign_scoring}
\end{figure*}

Bleualign \cite{Sennrich2010MTbasedSA,sennrich-volk-2011-iterative} uses MT and BLEU \cite{papineni-etal-2002-bleu} to align sentences. 
Even though BLEU has been criticised as a measure of translation quality and is not considered reliable on a sentence level \cite{callison-burch-etal-2006-evaluating}, the authors of Bleualign point out that judging the quality of a translation is harder than deciding whether two sentences are possible translations of each other. Furthermore, they find that BLEU is very sensitive to misalignments, indicating that it should be capable of discriminating between aligned and unaligned sentence pairs. BLEU is usually measured on up to 4-grams. Too often, for the purposes of sentence alignment, this yields a score of 0 
so Bleualign uses 2-grams.
Furthermore, when comparing two sentences, the BLEU scores are different depending on which of the sentences is the hypothesis, due to the brevity penalty in BLEU. Therefore, Bleualign translates both directions when possible and uses the mean as the final score. 
In the first pass of the alignment algorithm, a set of 1--1 beads are identified.
In the second pass, all unaligned sentences that fall between the beads, are extracted and a list generated of all possible 1-, 2- or 3-sentence sequences composed of the unaligned sentences and the beads. BLEU scores are then calculated for the Cartesian product of the two lists. If any 1--$n$ alignment scores higher than the bead, it is replaced in the graph and the step is repeated.

In Vecalign, \citet{thompson-koehn-2019-vecalign} use the similarity between sentence embeddings as the scoring function, employing LASER \cite{artetxe-schwenk-2019-massively} for scoring alignment candidates. In the alignment algorithm, recursive approximation is used to reduce the search space.

\begin{figure*}[t]
\centering
  \centering
  \includegraphics[height=0.15\textheight]{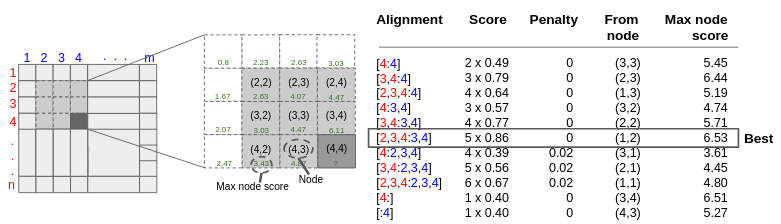}
  \caption{The maximum node score is calculated by adding the alignment score to the previously calculated maximum score of the node the alignment leads from. The LaBSE score is multiplied by the number of sentences comprising the alignment, e.g. alignment [2,3,4:3,4] has five sentences and thus the LaBSE score is multiplied by five. The max score for the node is found by adding the maximum score for node (1,2) to the alignment score.}
  \label{fig:sentalign_path}
\end{figure*}

\section{The SentAlign System}
\label{sentalign_system}


In this section, we present SentAlign, a highly accurate sentence aligner 
capable of evaluating all possible alignment paths through fairly large documents, using a LaBSE-based scoring mechanism. 
Our alignment approach is of quadratic complexity, $O(n^2)$, and in order to handle very large files, we apply a DaC approach. When the total nodes in the alignment graph exceed a user-defined maximum, by default set to $4,000,000$, the DaC-mechanism is activated in order to reduce the time complexity when aligning the documents.

The main components of the SentAlign system illustrated in Figure \ref{fig:sentalign_architecture} are the scoring mechanism, the alignment or pathfinding algorithm, a DaC-module to deal with very large files, and a readjustment module to compensate for shortcomings in the scoring mechanism.


\subsection{Scoring}
SentAlign uses LaBSE to score sentence-pair candidates. A minimum threshold score, defined by the user, is required for a sentence pair to be accepted. For each node $[i:j]$ in the alignment graph (where $i$ is a sentence in the source language and $j$ is a sentence in the target language), scores for all possible alignment combinations ending in that node are calculated. 
The user can set a maximum number of sentences that can be merged on either side of the alignment. If merging up to three sentences on each side is allowed, a total of $3\times3=9$ scores are compared for each node, as illustrated in Figure \ref{fig:sentalign_scoring}. If no alignment reaches the LaBSE threshold score, $S_{min}$, insertion and deletion functions are applied and the edges to the node obtain the score $S_{min}$. If the user wants to penalize long sentences, a user-defined maximum can be set for the number of words in either language. When either side of an alignment exceeds that maximum, a penalty is applied to the alignment score. 
The user can also define a maximum number of segment merges before a penalty is applied. That penalty is only applied in the pathfinding-phase (Section \ref{sec:pathfinding}) and not when readjusting the path (Section \ref{sec:readjusting}). This penalty is set in order to favour shorter alignments and to deter the aligner from merging multiple sentences in one alignment when it is possible to find multiple shorter alignments instead. 
SentAlign seeks a maximum score for a given node in the alignment graph, $S_{node}$, and finds it by adding the alignment scores to the score of the node they connect from after penalties are applied.

\subsection{Pathfinding}
\label{sec:pathfinding}
The alignment problem can be seen as a way of finding the optimal path through an ${N}\times{M}$ matrix, where $N$ and $M$ are the number of source and target sentences, respectively. 
As we allow for insertions, deletions and merges of multiple sentences on either side, we calculate the best path from the initial node $[0,0]$ to all other nodes in the graph using a version of Dijkstra's algorithm \cite{dijkstra1959note}.  Our objective is to maximize the score at each node, in contrast to the original algorithm, which minimizes scores. This allows for large missing parts of text in either language without straying from the right path.

After all possible alignment scores have been calculated for a given node, an alignment function is chosen. If none of the alignments reach $S_{min}$, insertion and deletion alignment functions are applied and $S_{min}$ assigned to the value of the resulting null alignments.
If one or more of the possible $n$--$m$ ($n\ge1$) alignments has a score above the $S_{min}$ threshold, we assign the alignment edge a value equalling the LaBSE score multiplied by the total number of sentences merged in both languages, and add penalty-adjustments to calculate the alignment score, as illustrated in Figure \ref{fig:sentalign_path}. Finally, we select the alignment obtaining in the highest score for $S_{node}$. 
This process is repeated for each node until node $(n, m)$ is reached. At that point, we have the optimal score from $(0, 0)$ to $(n, m)$ and mark the path by tracing backwards through the recorded edges.

\subsection{Divide and Conquer}
With more lines to align, the search space grows exponentially, affecting alignment speed.
\citet{Zhang2022ImproveSA} shows that for a quadratic time complexity sentence-alignment algorithm, chunking the parallel texts to be aligned using hard delimiters can reduce the time complexity to $O(n\log n)$.
SentAlign allows the user to define a threshold for dividing up the search space. If the search space is larger than the user-defined threshold allows, the tool searches for high-confidence alignments to use as hard delimiters for dividing the search space into multiple smaller chunks, 
$k+1$ chunks for $k$ hard delimiters. The aim is to find the minimum amount of alignments to use as hard delimiters to split the parallel texts into chunks of manageable size. 

SentAlign looks for 1--1 alignments in the middle half of the parallel texts to use as hard delimiters, with the middle half defined as the sentences in between the first and last 25\% of the sentences in the texts. One of two approaches is chosen, depending on the size of the files to align. The first choice is to employ the Gale--Church algorithm to align the parallel text/chunk under consideration, score the resulting 1--1 alignments using LaBSE and choose the highest-scoring alignment as a hard delimiter. If the parallel files are very large, running Gale--Church will take an excessive amount of time so SentAlign uses a fallback approach. When file size surpasses a second threshold, it resorts to a greedy algorithm that calculates LaBSE scores for 1--1 alignments in the allowed range and selects the highest one. 
When the hard-delimiter is found, the parallel text is split into two chunks. If the chunks are still too large, the process is repeated until all chunks of parallel text have the desired search space size.

\subsection{Readjusting the Path}
\label{sec:readjusting}
\citet{thompson-koehn-2019-vecalign} argue that sentence alignment should seek a minimal parallel pair, the pair having the fewest mergers while still being truly parallel. They find that dynamic programming with cosine similarity favours many-to-many alignments over 1--1 alignments, an effect we also find when using the scoring and alignment mechanism described above. To counteract this and produce more accurate alignments, SentAlign finishes by re-evaluating each alignment in the selected path by taking another look at mergers, insertions and deletions. 

First, SentAlign investigates all $n$$\times$$m$ alignments, where ($n>1$) and ($m>1$), and searches for the highest-scoring alignment which is a subset of the one being  investigated. If one is found that has a higher score than the original alignment, SentAlign amends the alignment path to add that as well as any other sentence pairs scoring above $S_{min}$. If any sentences are left they are added to the list of null alignments, containing previous insertions and deletions. 
Second, SentAlign looks at the list of non-aligned source and target sentences, i.e. null alignments. If a non-aligned sentence is adjacent to a sentence which has been aligned, SentAlign tries merging it to that alignment and calculates the LaBSE score. If the score increases, the path is amended. This is repeated until no more amendments can be made.

When the re-evaluation is finished, SentAlign writes out the set of alignments generated by the selected path through the alignment graph.

\section{Evaluation}
\label{sec:Evaluation}
We evaluated SentAlign by comparing the system to other sentence aligners, both using sentence alignment evaluation sets and by testing the impact on downstream MT task.

\subsection{Two evaluation sets}
\label{sec:two_evaluation_sets}
We compared SentAlign to five other sentence aligners: Vecalign, Bleualign, Gargantua, Hunalign and Gale-Church (using their default settings). We used two evaluation sets:
\begin{enumerate}
    \item The manually aligned German--French evaluation set created from the Text+Berg corpus \cite{volk-etal-2010-challenges}, first used to evaluate Bleualign and commonly used for sentence alignment evaluation since then. 
    \item We compiled an evaluation set for English--Icelandic sentence alignment from 10 aligned documents in five subcorpora of the ParIce corpus \cite{barkarson-steingrimsson-2019-compiling}. The evaluation set \cite{20.500.12537/150} is available under an open licence and contains a total of 549 sentence alignments.\footnote{\url{http://hdl.handle.net/20.500.12537/150}} These documents are arguably easier to align than the Text+Berg documents, as none of them contain long stretches of non-alignments and there are few $n$--$m$ merging alignments. 
\end{enumerate}
When translating the evaluation sets for Bleualign, we use OPUS-MT\footnote{\url{https://opus.nlpl.eu/Opus-MT/}} \cite{tiedemann-thottingal-2020-opus}.

We used the development set from the Text+Berg corpus to search for the best parameters for SentAlign. We found the best $S_{min}$ (LaBSE) threshold to be $0.4$, maximum number of words per language before applying a length penalty to be $80$, and the penalty for each word exceeding that maximum to be $0.01$. We performed a complete search through the alignment matrix, without chunking the search space by finding anchors as all the evaluation files were within the limits for the hard delimiters.

While none of the aligners used, with the exception of Bleualign, allow reordering of sentences in cases of possible crossing alignments, there are examples of such alignments in the Text+Berg evaluation set, which makes it impossible for other aligners to attain a perfect score. Furthermore, a few entries of null alignments are missing from the files distributed with Bleualign. To maintain consistency with previous reported scores, we did not make any changes to the evaluation set. As only some null alignments are included in the evaluation set and some are not, the results can be different based on whether a given sentence aligner returns null alignments or only useful alignments. We thus only calculated precision on non-null alignments, i.e.\ alignments that are true sentence pairs. 

Following the original Bleualign paper, in Table \ref{tab:textberg_eval} we report results both under the strict condition where exact matches between the gold alignment and the hypothesis are demanded, and under the lax condition where a hypothesis is true if there is an overlap with a gold alignment on both language sides. Under the lax condition, a 2--2 alignment, which is recognized as two 1--1 alignments, will yield two true positives, while it would yield two false positives under the strict condition.

\begin{table}\small
\begin{center}
\begin{tabular}{|l|r|r|r||r|r|r|}
\hline
\multicolumn{7}{|c|}{\textbf{Alignment results on Text+Berg}} \\
\hline
 &\multicolumn{3}{c||}{Strict}&\multicolumn{3}{c|}{Lax} \\
Algorithm &P&R&$F_1$&P&R&$F_1$ \\
\hline
Gargantua & 0.76 & 0.75 & 0.76 &0.89&0.78&0.83 \\
Hunalign & 0.66 & 0.69 & 0.67 &0.86&0.74&0.80 \\
Gale--Ch. & 0.68 & 0.69 & 0.69 &0.80&0.73&0.76 \\
Vecalign & 0.90 & 0.90 & 0.90 &0.99&0.91&0.95 \\
Bleualign & 0.93 & 0.66 & 0.77 & \textbf{1.00} & 0.68 & 0.81  \\
SentAlign & \textbf{0.94} & \textbf{0.93} & \textbf{0.93} &\textbf{1.00}&\textbf{0.93}&\textbf{0.96} \\
\hline
\end{tabular}
\end{center}
\caption{Evaluating on the German--French Text+Berg evaluation set. The highest scores are in bold. SentAlign outperforms all systems both for the strict and lax conditions, although Bleualign has a perfect score for precision, just like SentAlign.}
\label{tab:textberg_eval}
\end{table}

We use the same settings and parameters as before for all the aligners when we evaluate on the English--Icelandic evaluation set. 
As with the evaluation set from Text+Berg, the sentence embeddings-based alignment systems SentAlign and Vecalign are the most accurate. Using this evaluation set, the scores are higher for all aligners (see Table \ref{tab:sentalign_isl_eval}). Even though we are missing a development set for the en--is language pair and used the SentAlign parameters set for the Text+Berg de--fr development set, SentAlign does well. The results might possibly improve even more if we were to search for the best values for this sort of en--is data as the acceptance threshold for LaBSE may be different for different language pairs. While we found that $0.4$ was the optimum threshold score for the Text+Berg corpus, \citet{feng-etal-2022-language} set their threshold when mining sentences from CommonCrawl to $0.6$. This suggests that analysis of the languages to be processed could be useful on a case-by-case basis. 

\begin{table}\small
\begin{center}
\begin{tabular}{|l|r|r|r||r|r|r|}
\hline
\multicolumn{7}{|c|}{\textbf{Alignment results on English--Icelandic evaluation set}} \\
\hline
 &\multicolumn{3}{c||}{Strict}&\multicolumn{3}{c|}{Lax} \\
Algorithm &P&R&$F_1$&P&R&$F_1$ \\
\hline
Gargantua &0.82&0.76&0.79&0.89&0.78&0.83 \\
Hunalign &0.72&0.75&0.73&0.87&0.78&0.82 \\
Gale--Ch. &0.78&0.79&0.79&0.87&0.81&0.84\\
Vecalign &0.92&0.94&0.93&0.97&0.95&0.96 \\
Bleualign &0.93&0.78&0.85&0.98&0.79&0.88 \\
SentAlign &\textbf{0.95}&\textbf{0.95}&\textbf{0.95}&\textbf{0.99}&\textbf{0.96}&\textbf{0.97} \\
\hline
\end{tabular}
\end{center}
\caption{Evaluating on the English--Icelandic evaluation set. The highest scores are in bold. SentAlign outperforms other systems and Vecalign is the only other aligner that comes close.}
\label{tab:sentalign_isl_eval}
\end{table}

\subsection{Downstream MT}

For the downstream MT task, we aligned English and Icelandic documents containing EEA regulations and directives. These documents are available as a subcorpus of ParIce 21.10\footnote{\url{http://hdl.handle.net/20.500.12537/145}} which is published with an evaluation set in that domain.\footnote{\url{http://hdl.handle.net/20.500.12537/146}}
We used fairseq \cite{ott-etal-2019-fairseq} to train Transformer$_\mathrm{BASE}$ models \cite{vaswani2017}, and  SacreBleu \cite{post-2018-call} to calculate BLEU scores and statistical significance using the pairwise bootstrap test \cite{koehn-2004-statistical}. Table \ref{tab:alignment_eval_bleu} reports the results for all systems,
showing that SentAlign achieved the best results of the six aligners evaluated, with BLEU scores of $42.8$ and $53.6$, for en$\rightarrow$is and is$\rightarrow$en, respectively. A significance test shows that this is significantly better than all the other aligners.

\begin{table}\small
\begin{center}
\begin{tabular}{|l|r|r|r|}
\hline
\multicolumn{4}{|c|}{\textbf{Downstream MT Task}} \\
\hline
 \textbf{Sentence Aligner}&no. pairs&en$\rightarrow$is&is$\rightarrow$en \\
\hline
Gargantua &606,768&39.1&48.9 \\
Hunalign &717,879&41.4&52.1 \\
Gale--Church &683,813&41.8&51.4  \\
Vecalign &670,595&41.8&51.7 \\
Bleualign &627,019&42.0&53.0 \\
SentAlign &877,485&\textbf{\textit{42.8}}&\textbf{\textit{53.6}} \\
\hline
\end{tabular}
\end{center}
\caption{Results for MT systems trained on sentence pairs generated by different alignment tools. The table shows number of aligned pairs generated by the tools and BLEU scores for the MT systems. Bold and italic scores are the highest scores for each category and significantly higher than other systems.}
\label{tab:alignment_eval_bleu}
\end{table}

\section{Conclusion}
SentAlign is an accurate, scalable and easy-to-use sentence alignment system. It uses the LaBSE model, which has been trained to generate sentence embeddings in 109 languages, to score alignment candidates. The alignment algorithm considers all possible paths through the alignment graph where the number of merges for adjoining sentences in each language is under a user-set threshold, and the maximum number of nodes in the search space is less than the DaC-threshold. Evaluation on two sentence alignment evaluation sets, as well as on a downstream MT task, show that the aligner is highly competitive, outperforming other alignment systems in most regards. SentAlign is distributed under an Apache 2.0 licence.

\section*{Limitations}
SentAlign can deliver accurate results for medium to high-resource languages in common scenarios. 
It is capable of evaluating all possible alignment paths through the alignment graph for parallel documents. However, as the documents get larger this may be at the cost of speed and, for very large documents, alignment time would be too long for practical use. To address this, our DaC-mechanism is applied, which enables the alignment of very large documents within reasonable time limits. Nevertheless, we can expect the system to run into problems when the number of lines in each document reaches multiple tens of thousands, due to memory constraints as well as the time factor.

LaBSE is trained on 109 languages. As noted in Section \ref{sec:two_evaluation_sets}, the optimal minimum score threshold may be different between language pairs, impacting insertions and deletion made by the aligner. Furthermore, we can expect the accuracy of our scoring function to fall if the tool is used on languages not represented in the LaBSE training data.

Finally, we used the default OPUS-MT models for aligning with Bleualign. By replacing the OPUS-MT models with higher quality models, the results for Bleualign may be further improved.


\section*{Acknowledgements}
This work was supported by the The Icelandic Centre for Research, RANNIS grant number 228654-051, and by the ADAPT Centre for Digital Content Technology which is funded under the Science Foundation Ireland (SFI) Research Centres Programme (Grant No. 13/RC/2106) and is co-funded under the European Regional Development Fund.

\bibliography{emnlp2023}
\bibliographystyle{acl_natbib}
\end{document}